\documentclass[11pt,a4paper]{article}
\usepackage[hyperref]{acl2019}
\usepackage{times}
\usepackage{latexsym}
\usepackage{bm}
\usepackage{multirow}
\usepackage{amsfonts}
\usepackage{amsmath}
\usepackage{booktabs}
\usepackage{xspace}
\usepackage{url}

\aclfinalcopy 

\newcommand\BERTBASE{BERT$_{\small \textsc{BASE}}$\xspace}
\newcommand\BERTLARGE{BERT$_{\small \textsc{LARGE}}$\xspace}
\newcommand\OCNBASE{OCN$_{\small \textsc{BASE}}$\xspace}
\newcommand\OCNLARGE{OCN$_{\small \textsc{LARGE}}$\xspace}

\title{Option Comparison Network for Multiple-choice Reading Comprehension}

\author{Qiu Ran\thanks{\ \ indicates equal contribution}, Peng Li\footnotemark[1], Weiwei Hu, Jie Zhou \\
  Pattern Recognition Center, WeChat AI, Tencent Inc, China \\
  \texttt{\{soulcaptran,patrickpli,weiweihu,withtomzhou\}@tencent.com}}

\date{}

\begin{document}
\maketitle
\begin{abstract}
Multiple-choice reading comprehension (MCRC) is the task of selecting the correct answer from multiple options given a question and an article. Existing MCRC models typically either read each option independently or compute a fixed-length representation for each option before comparing them. However, humans typically compare the options at multiple-granularity level before reading the article in detail to make reasoning more efficient. Mimicking humans, we propose an option comparison network (OCN) for MCRC which compares options at word-level to better identify their correlations to help reasoning. Specially, each option is encoded into a vector sequence using a skimmer to retain fine-grained information as much as possible. An attention mechanism is leveraged to compare these sequences vector-by-vector to identify more subtle correlations between options, which is potentially valuable for reasoning. Experimental results on the human English exam MCRC dataset RACE show that our model outperforms existing methods significantly. Moreover, it is also the first model that surpasses Amazon Mechanical Turker performance on the whole dataset.
\end{abstract}

\begin{table}
    \small
    \centering
    \begin{tabular}{|p{0.95\columnwidth}|}
    \hline
    {\bf Article:} \\
    Are you a crazy chocolate fan? Have you heard about Hershey's Kisses? Do you love the movie Charlie and the Chocolate Factory? If your answer was, "yes", to any of the questions, then my experience will make you jealous. I just went to the famous Hershey Chocolate Factory! ...... When we arrived at the factory, we realized that this was much more than just a factory. The whole town is a chocolate-themed amusement park ......
    Jason, our tour guide, began telling us about this quiet little town ......
    Jason went on, ``The factory first started on a small farm. It developed very fast. So they built this town for factory workers to live in. Then they built hotels, hospitals, stadiums, theaters and even museums with the theme of chocolate. Isn't that cool?''
    ``Yes, a hundred times yes!'' I yelled ( ) with delight.\\
    {\bf Question:}\\
    What can we know from the writer's answer to the guide?\\
    {\bf Options:}\\
    A. The writer had never heard about Hershey Chocolate. \\
    B. The writer didn't want to visit the factory any more. \\
    C. The writer had visited the factory before. \\
    D. The writer couldn't wait to visit the factory. \\
    {\bf Answer:} D\\
    \hline
    \end{tabular}
    \caption{An MCRC example from the RACE dataset.}
    \label{tab:example}
\end{table}

\section{Introduction}
\label{sec:intro}

Multiple-choice reading comprehension (MCRC) aims to selecting the correct answer from a set of options given a question and an article. As MCRC requires both understanding of natural language and world knowledge to distinguish correct answers from distracting options, it is challenging for machine and a good testbed for artificial intelligence. With the rapid development of deep learning, various neural models have been proposed for MCRC and achieve promising results in recent years~\cite{StanfordAR,yin2016HCQA,trischler2016,GAReader,MRU,ElimiNet,HAF,co-matching,DFN,Reading-Strategies-Model,shuailiang2019DCMN}.

Comparing options before reading the article in detail is a commonly used strategy for humans when solving MCRC problems. By comparing the options, the correlations between the options can be identified and people only need to pay attention to the information related to the correlations when reading the article. As a result, questions can be answered more efficiently and effectively. Taking Table~\ref{tab:example} as an example, by comparing option B and D, people may identify that the key difference is whether the writer would like to visit the factory, which can be decided easily by skimming the article.

However, the strategy is not adopted by most existing MCRC methods.
The Stanford AR~\cite{StanfordAR} and GA Reader~\cite{GAReader} variants used in~\cite{RACE} encode question and article independent of options, ignoring their correlations.
In contrast, \newcite{co-matching} and \newcite{shuailiang2019DCMN} leverage sophisticated matching mechanisms to gather the correlation information, while \newcite{Reading-Strategies-Model} relies on a pre-trained language model~\cite{OpenAIGPT} to extract such information. 
Nevertheless, none of them consider the correlations between options explicitly.
To the best of our knowledge, \cite{HAF} is the only work that considers option correlations explicitly. Whereas, the options are compressed into fixed-length vectors before being compared, which may make it hard for a model to identify subtle differences or similarities between options.

To gather option correlation information more effectively, we propose option comparison network (OCN), a novel method for MCRC which explicitly compares options at word-level to mimic the aforementioned human strategy. Specially, we first use a skimmer network to encode options into vector sequences independently as their features. Then for each option, it is compared with other options {\em one-by-one} at {\em word-level} using an attention-based mechanism in vector space to identify their correlations. Finally, the article is reread with the gathered correlation information to do reasoning and select the correct answer.
As options are compared one-by-one, the correlations between each pair of options can be explicitly identified. By comparing options at word-level, we allow the model to detect subtle correlations more easily.

With a BERT~\cite{BERT} based skimmer, our method outperforms the state-of-the-art baselines with large margins on RACE, a human exam MCRC dataset created by experts for assessing the reading comprehension skills of students, indicating the effectiveness of our model. More importantly, it is the first time that a model surpasses the Amazon Mechanical Turker performance on this dataset.

\section{Option Comparison Network}

Suppose we have a question $Q$ with $n$ tokens $\{w^q_1,w^q_2,\cdots,w^q_n\}$, an article $P$ with $m$ tokens $\{w^p_1,w^p_2,\cdots,w^p_m\}$, and a candidate answer set $\mathcal{O}$ with $K$ options $\{O_1, O_2,\cdots,O_K\}$. Each option $O_k$ consists of $n_k$ tokens $\{w_1^o,w_2^o,\cdots,w_{n_k}^o\}$.
Formally, MCRC is to select the correct answer $\hat{O}$ from the candidate answer set $\mathcal{O}$ given question $Q$ and article $P$.

Our model selects the correct answer from the candidate answer set in four stages. First, we concatenate each (article, question, option) triple into a sequence and use a skimmer to encode them into vector sequences (Sec.~\ref{sec:encoding}). Then an attention-based mechanism is leveraged to compare the options (Sec.~\ref{sec:comparison}). Next the article is reread with the correlation information gathered in last stage as extra input (Sec.~\ref{sec:rereading}). And finally the probabilities for each option to be the correct answer are computed (Sec.~\ref{sec:prediction}). The details will be introduced in the following sections.

\subsection{Option Feature Extraction}
\label{sec:encoding}
A skimmer network is used to skim the options independently together with the question and article to extract option features.
As BERT~\cite{BERT} has been shown to be a powerful feature extractor for various tasks, it is used as the skimmer. Specially, for option $O_k$, it is concatenated with the question $Q$ and article $P$, denoted as $\langle P;Q;O_k\rangle$~\footnote{Delimiter [SEP] are added between $P$, $Q$ and $O_k$. We omit [SEP] from the notation for brevity.}. Then the sequence is fed to BERT to compute their vector space encoding, which is denoted as 
\begin{equation}
    [\bm{P}^{enc};\bm{Q}^{enc};\bm{O}^{enc}_k]=\mathrm{BERT}\left(\langle P;Q;O_k\rangle\right)
\end{equation}
where $\bm{P}^{enc} \in \mathbb{R}^{d\times m}$, $\bm{Q}^{enc} \in \mathbb{R}^{d\times n}$, $\bm{O}^{enc}_k \in \mathbb{R}^{d\times n_k}$, and $\mathrm{BERT}(\cdot)$ denotes the network defined in~\cite{BERT} \footnote{We refer the readers to~\cite{BERT} for details of $\mathrm{BERT}(\cdot)$.}. 

As question and options are closely related, we use
\begin{equation}
    \bm{O}_k^q=[\bm{Q}^{enc}|\bm{O}^{enc}_k] \in \mathbb{R}^{d\times n_k'}
\end{equation}
as features of $O_k$, where $n_k'=n + n_k$ and $[\cdot|\cdot]$ denotes row-wise concatenation.

\subsection{Option Correlation Features Extraction}
\label{sec:comparison}
This module is used to compare options at word level to extract option correlation information to support reasoning. For each option, an attention-based mechanism is used to compare it with all the other options to gather the correlation information.

Given input matrices $\bm{U}\in \mathbb{R}^{d\times N}$ and $\bm{V}\in \mathbb{R}^{d\times M}$, the attention weight function $\texttt{Att}(\cdot)$ specified by the parameter $\bm{v} \in \mathbb{R}^{3d}$ is defined as
\begin{eqnarray}
    s_{ij} & = &\bm{v}^{\mathrm{T}}\left[\bm{U}_{:i};\bm{V}_{:j};\bm{U}_{:i} \circ \bm{V}_{:j}\right]\label{eqn:sim}\\
	\bm{A} & = & \texttt{Att}\left(\bm{U}, \bm{V};\bm{v}\right) \\
		   & = & \left[\frac{\exp(s_{ij})}{\sum_i\exp(s_{ij})}\right]_{i,j}
\end{eqnarray}
where $[\cdot;\cdot]$ denotes column-wise concatenation, $\circ$ denotes the element-wise multiplication operation, and $\bm{A}\in \mathbb{R}^{N\times M}$ is the attention weight matrix.

The option correlation features are extracted in three steps as follows:

First, an option is compared with all other options one-by-one to collect the pairwise correlation information. Specially, for option $O_k$, the information $\widetilde{\bm{O}}_k^{(l)}\in \mathbb{R}^{2d \times n_k'}$ gathered from option $O_l$ is computed as
\begin{eqnarray}
	\bar{\bm{O}}_{k}^{(l)}&=&\bm{O}^q_l \texttt{Att}(\bm{O}^q_l,\bm{O}^q_k;\bm{v}_o) \\
	\widetilde{\bm{O}}_k^{(l)}&=&\left[\bm{O}^q_k-\bar{\bm{O}}_{k}^{(l)}; \bm{O}^q_k\circ\bar{\bm{O}}_{k}^{(l)}\right]
\end{eqnarray}

Then the pairwise correlation information gathered for each option is fused to get the option-wise correlation information, which is defined as
\begin{equation}
    \widetilde{\bm{O}}^c_k=\tanh\left(\bm{W}_c\left[\bm{O}^q_k; \left\{\widetilde{\bm{O}}_k^{(l)}\right\}_{l\neq k}\right] + \bm{b}_c\right)
\end{equation}
where $\bm{W}_c\in\mathbb{R}^{d\times (d+2d(|O|-1))}$ and $\bm{b}_c\in\mathbb{R}^{d}$. Note that option $O_k$ is not compared with itself. 

Finally, an element-wise gating mechanism is leveraged to fuse the option features with the option-wise correlation information to produce the option correlation features $\bm{O}^c_{k}$. Specially, the gates $\bm{g}_k\in\mathbb{R}^{d\times n_k'}$ are defined as
\begin{equation}
  \bm{g}_{k,:i}=\mathrm{sigmoid}\left(\bm{W}_g[\bm{O}^q_{k,:i};\widetilde{\bm{O}}^c_{k,:i}; \widetilde{\bm{Q}}] + \bm{b}_g\right)
\end{equation}
where $\bm{g}_{k,:i}$ denotes the $i$-th column of $\bm{g}$, and $\widetilde{\bm{Q}}\in\mathbb{R}^d$ is the attentive-pooling of $\bm{Q}^{enc}$ defined as
\begin{gather}
\bm{A}^q=\mathrm{softmax}\left(\bm{v}_a^{\mathrm{T}}\bm{Q}^{enc}\right)^{\mathrm{T}}, \bm{v}_a \in \mathbb{R}^d\\
\widetilde{\bm{Q}}=\bm{Q}^{enc} \bm{A}^q
\end{gather}
The option correlation features $\bm{O}^c_{k}\in\mathbb{R}^{d\times n_k'}$ are computed as
\begin{equation}
  \bm{O}^c_{k,:i}=\bm{g}_{k,:i} \circ \bm{O}^q_{k,:i} + (1-\bm{g}_{k,:i}) \circ \widetilde{\bm{O}}^c_{k,:i}
\end{equation}
Note that $\bm{O}^c_{k}$ is not compressed into a fixed-length vector, because we believe this will enable our model to utilize the correlation information in a more flexible way.

\subsection{Article Rereading}
\label{sec:rereading}
Mimicking humans, the article will be reread with the option correlation features as extra input to gain deeper understanding. Specially, the co-attention~\cite{DCN} and self-attention~\cite{r-net} mechanisms are adopted for rereading.
First, for each option $O_k$, co-attention is performed as
\begin{eqnarray}
	\bm{A}^c_k &=& \texttt{Att}\left(\bm{O}^c_k, \bm{P}^{enc};\bm{v}_p\right) \in \mathbb{R}^{n_k'\times m} \\
	\bm{A}^p_k &=& \texttt{Att}\left(\bm{P}^{enc}, \bm{O}^c_k;\bm{v}_p\right) \in \mathbb{R}^{m\times n_k'} \\
	\hat{\bm{O}}^p_k&=&[\bm{P}^{enc};\bm{O}^c_k\bm{A}^c_k]\bm{A}^p_k \in \mathbb{R}^{2d\times n_k'}
\end{eqnarray}
Then $\hat{\bm{O}}^p_k$ is fused with option correlation features $\bm{O}^c_k$ as
\begin{equation}
    \widetilde{\bm{O}}^p_k=\mathrm{ReLU}(\bm{W}_p[\bm{O}^c_k;\hat{\bm{O}}^p_k]+\bm{b}_p)
\end{equation}
where $\widetilde{\bm{O}}^p_k\in \mathbb{R}^{d\times n_k'}$, $\bm{W}_p\in\mathbb{R}^{d\times 3d}$, and $\bm{b}_p\in\mathbb{R}^{d}$.
Finally, the full-info option representation $\bm{O}^f_k\in \mathbb{R}^{d\times n_k'}$ for option $O_k$ is computed with self-attention as
\begin{eqnarray}
    \widetilde{\bm{O}}^s_k&=&\widetilde{\bm{O}}^p_k\texttt{Att}(\widetilde{\bm{O}}^p_k, \widetilde{\bm{O}}^p_k;\bm{v}_r) \\
    \widetilde{\bm{O}}^f_k&=&[\widetilde{\bm{O}}^p_k; \widetilde{\bm{O}}^s_k; \widetilde{\bm{O}}^p_k-\widetilde{\bm{O}}^s_k;\widetilde{\bm{O}}^p_k\circ\widetilde{\bm{O}}^s_k]\\
    \bm{O}^f_k&=&\mathrm{ReLU}(\bm{W}_f\widetilde{\bm{O}}^f_k+\bm{b}_f)
\end{eqnarray}
where $\bm{W}_f\in\mathbb{R}^{d\times 4d}$ and $\bm{b}_f\in\mathbb{R}^{d}$.

\subsection{Answer Prediction}
\label{sec:prediction}
The score $s_k$ of option $O_k$ to be the correct answer is computed as
\begin{equation}
	s_k=\bm{v}_s^\mathrm{T}\texttt{MaxPooling}\left(\bm{O}^f_k\right)
\end{equation}
where $\texttt{MaxPooling}(\cdot)$ performs row-wise max pooling and $\bm{v}_s\in\mathbb{R}^{d}$.

The probability $P(k|Q,P,O)$ of option $O_k$ to be the correct answer is computed as
\begin{equation}
    P(k|Q,P,\mathcal{O})=\frac{\exp(s_k)}{\sum_i\exp(s_i)}
\end{equation}
And the loss function is defined as
\begin{equation}
    J(\theta)=-\frac{1}{N}\sum_i\log(P(\hat{k}_i|Q_i,P_i,\mathcal{O}_i)) + \lambda ||\theta||^2_2
\end{equation}
where $\theta$ denotes all trainable parameters, $N$ is the training example number, and $\hat{k}_i$ is the ground truth for the $i$-th example.

\begin{table*}
    \vspace{-1em}
	\centering
	\small
	\begin{tabular}{lcccc}
		\toprule
		Model & Pre-training & RACE-M & RACE-H & RACE \\
		\midrule
		\multicolumn{5}{c}{Single Model} \\
		\midrule
		Stanford AR~\cite{StanfordAR} & / & 44.2 & 43.0 & 43.3 \\
		GA Reader~\cite{GAReader} & / & 43.7 & 44.2 & 44.1 \\
		ElimiNet~\cite{ElimiNet} & / & 44.4 & 44.5 & 44.5 \\ 
		HAF~\cite{HAF} & / & 45.0 & 46.4 & 46.0 \\ 
		Hier-Co-Matching~\cite{co-matching} & / & 55.8 & 48.2 & 50.4 \\ 
		DFN~\cite{DFN} & / & 51.5 & 45.7 & 47.4 \\  
		MRU~\cite{MRU} & / & 57.7 & 47.4 & 50.4 \\ 
		OpenAI GPT~\cite{OpenAIGPT} & GPT & 62.9 & 57.4 & 59.0 \\ 
		Reading Strategies Model~\cite{Reading-Strategies-Model} & GPT & 69.2 & 61.5 & 63.8 \\ 
		DCMN~\cite{shuailiang2019DCMN} & BERT & {\bf 76.7} & 68.5 & 70.9 \\
		\BERTBASE & BERT & 70.5 & 63.0 & 65.2 \\
		\BERTLARGE & BERT & 76.4 & 68.8 & 71.0 \\
		\OCNBASE & BERT & 71.6 & 64.8 & 66.8 \\
		\OCNLARGE & BERT & {\bf 76.7} & {\bf \underline{69.6}} & {\bf 71.7} \\
		\midrule
		\multicolumn{5}{c}{Ensemble} \\
		\midrule
		GA Reader~\cite{GAReader} & / & / & / & 45.9 \\
		ElimiNet~\cite{ElimiNet} & / & 47.7 & 46.1 & 46.5 \\ 
		DFN~\cite{DFN} & / & 55.6 & 49.4 & 51.2 \\ 
		MRU~\cite{MRU} & / & 60.2 & 50.3 & 53.3 \\ 
		Reading Strategies Model~\cite{Reading-Strategies-Model} & BERT & 72.0 & 64.5 & 66.7 \\ 
		DCMN~\cite{shuailiang2019DCMN} & BERT & 77.9 & \underline{69.8} & 72.1 \\
		\OCNBASE & BERT & 74.4 & 67.0 & 69.2 \\
		\OCNLARGE & BERT & {\bf 78.4} & {\bf \underline{71.5}} & {\bf \underline{73.5}} \\
		\midrule
		Amazon Mechanical Turker & / & 85.1 & 69.4 & 73.3 \\
		Human Ceiling Performance & / & 95.4 & 94.2 & 94.5 \\
		\bottomrule
	\end{tabular}
	\caption{Experimental results. The best results in each group are in bold, and those better than Amazon Mechanical Turker are underlined.}
	\label{tab:results}
	\vspace{-1em}
\end{table*}

\section{Experiments}
\subsection{Dataset}


We evaluate our model on RACE~\cite{RACE}, an MCRC dataset collected from the English exams for middle and high school students in China. The dataset is further devided into RACE-M and RACE-H, containing only data from middle school and high school examinations respectively. 
As the articles, questions and options are generated by English instructors for assessing the reading comprehension skills of humans, the dataset is inherently more difficult than other widely used reading comprehension datasets such as SQuAD~\cite{squad}. Analysis conducted in~\cite{RACE} shows that $59.2\%$ of the questions in RACE require reasoning, which is significantly higher than that of SQuAD ($20.5\%$). And the most frequent reasoning skills required are detail reasoning, whole-picture understanding, passage summarization, attitude analysis and world knowledge. Therefore, RACE is extremely challenging for MCRC models.

\subsection{Training Details}

Adam optimizer~\cite{adam} is used to train our model. 
The model is trained for 3 epochs with batch size 12 and learning rate $3\times 10^{-5}$ when \BERTBASE is used as the skimmer, and trained for 5 epochs with batch size 24 and learning rate $1.5\times 10^{-5}$ when \BERTLARGE is used.
For both cases, the learning rate linearly increases from 0.0 to the aforementioned value in the first $10\%$ training steps and then linearly decays until training is completed. The L2 weight decay $\lambda$ is set to 0.01. Articles, questions and options are trimmed to 400, 30 and 16 tokens respectively for memory and speed consideration.

\subsection{Experimental Results}
We compare our model with various state-of-the-art methods and the results are shown in Table~\ref{tab:results}, where \OCNBASE and \OCNLARGE denote our model with \BERTBASE and \BERTLARGE as the skimmer (Sec.~\ref{sec:encoding}) respectively.
From the results we can observe that:
(1) Our model outperforms the baselines significantly, indicating the effectiveness of our model.
(2) Our ensemble model with \BERTLARGE as the skimmer surpasses Amazon Mechanical Turker on the whole dataset and the margin on the RACE-H subset is significantly large. Moreover, \OCNLARGE also outperforms Amazon Mechanical Turker without ensembling. All these results indicate that our model has learned certain reasoning skills. 
(3) There is still a large gap between human ceiling performance and our model's performance. We believe this is because our model still struggles in complex reasoning as expected.
(4) All the models using pre-trained contextualized representations (GPT~\cite{OpenAIGPT} and BERT) outperform the other models with significantly large margins, indicating pre-training is a promising research direction for learning semantics from unsupervised data.

\begin{table}
	\centering
	\small
	\begin{tabular}{lccc}
		\toprule
		Model & RACE-M & RACE-H & RACE \\
		\midrule
		Ours (\BERTBASE) & 71.6 & 64.8 & 66.8 \\
		$\quad-$w/o Opt. Comp. & 71.5 & 63.9 &  66.1\\
		$\quad-$w/ ELMo & 50.9 & 45.7 & 47.2 \\
		\bottomrule
	\end{tabular}
	\caption{Ablation study. ``Opt. Comp.'' denotes option comparison.}
	\label{tab:ablation}
	\vspace{-1.1em}
\end{table}

The ablation study results are shown in Table~\ref{tab:ablation}. Removing the option comparison component (Sec.~\ref{sec:comparison}) causes significant performance drop, especially on RACE-H, indicating the effectiveness of considering the correlations between options. The performance of our model drops seriously when BERT is replaced with ELMo~\cite{elmo}, suggesting that BERT is a powerful feature extractor that can capture rich semantics.

\section{Conclusion and Future Work}
To leverage option correlations to improve reasoning ability, we propose option comparison network (OCN) for multiple-choice reading comprehension in this work. By representing options as vector sequences and comparing them vector-by-vector, we allow our model to identify the correlations between options more effectively.
Experimental results show that our model outperforms the state-of-the-art baselines significantly and surpasses Amazon Mechanical Turker on the whole RACE dataset for the first time, indicating that our model is effective and has learned certain reasoning skills.

As shown in the ablation study, our model relies on the pre-trained BERT model heavily. However, BERT model is large and slow. How to reduce the model size and improve its speed with acceptable performance drop is an interesting future work.

\bibliography{race-bert}
\bibliographystyle{acl_natbib}

\appendix
\end{document}